\documentclass[10pt,twocolumn,letterpaper]{article}

\usepackage{cvpr}
\usepackage{times}
\usepackage{epsfig}
\usepackage{graphicx}
\usepackage{amsmath}
\usepackage{amssymb}
\usepackage{lscape}

\usepackage[pagebackref=true,breaklinks=true,letterpaper=true,colorlinks,bookmarks=false]{hyperref}

\cvprfinalcopy 

\def\cvprPaperID{3676} 

\ifcvprfinal\fi
\begin{document}

\title{Boosted Cascaded Convnets for Multilabel Classification of Thoracic Diseases in Chest Radiographs}

\author{Pulkit Kumar\textsuperscript{*}\\\
Paralleldots, Inc.\\
{\tt\small pulkit@paralleldots.com}
\and
Monika Grewal\thanks{Authors contributed equally.}\\
Paralleldots, Inc.\\
{\tt\small monika@paralleldots.com}
\and
Muktabh Mayank Srivastava\\
Paralleldots, Inc.\\
{\tt\small muktabh@paralleldots.com}
}

\maketitle
\begin{abstract}
Chest X-ray is one of the most accessible medical imaging technique for diagnosis of multiple diseases. With the availability of ChestX-ray14, which is a massive dataset of chest X-ray images and provides annotations for 14 thoracic diseases; it is possible to train Deep Convolutional Neural Networks (DCNN) to build Computer Aided Diagnosis (CAD) systems. In this work, we experiment a set of deep learning models and present a cascaded deep neural network that can diagnose all 14 pathologies better than the baseline and is competitive with other published methods. Our work provides the quantitative results to answer following research questions for the dataset: 1) What loss functions to use for training DCNN from scratch on ChestX-ray14 dataset that demonstrates high class imbalance and label co occurrence? 2) How to use cascading to model label dependency and to improve accuracy of the deep learning model?
\end{abstract}

\section{Introduction}

Computer Aided Diagnosis (CAD) has been a well sought research field ever since the inception of medical imaging techniques and the interest is increasing with the advent of sophisticated medical imaging techniques. Among the existing medical imaging techniques, X-ray imaging is most commonly used technique for screening and diagnosis of lung related diseases e.g. pneumonia, cardiomegaly, lung nodules etc. The cost effectiveness of X-rays makes it most accessible diagnostic method for chest diseases in third world countries. The diagnosis of disease from a radiograph is, however, a time-consuming and challenging task. Thus, development of a CAD system for evaluation would increase the productivity of physicians and accessibility of better healthcare services in remote areas.

Recent years have observed significant rise in use of deep learning methods for analysis of medical imaging datasets. Deep learning methods have been rigorously applied for disease classification and image segmentation tasks. The gradual rise of interest in deep learning based diagnostic solutions can be attributed to their tremendous potential in modelling complex relationships between input and output variables and faster inference.  Training highly accurate deep learning systems in any domain requires a large annotated dataset. We only had relatively smaller datasets available for chest X-ray diagnosis until recently. A significant step in this direction is the work by Wang et. al.\cite{Wang2017ChestX-ray8:Diseases}, who developed a large dataset of chest X-ray images along with annotations for chest diseases through Natural Language Processing (NLP). The dataset is the largest open dataset for chest X-ray images, and therefore paves a path for development of better algorithms for automated diagnosis of chest diseases. 

Chest X-ray 14 dataset provides labels for 14 lung related diseases for 112,120 radiograph images of 1024 by 1024 resolution with above 90\% label accuracy. Most of the radiographs are labelled with more than one disease making it a typical multi-label classification problem. Apart from the inherent challenges posed by multi-label classification, the dataset has heavy imbalance in number of instances of individual classes. Moreover, the co-occurrence instances of each class with every other class are very high. For instance, the disease ‘Cardiomegaly’ co-occurs with disease ‘Effusion’ in 1060 images, whereas the total images of the diseases are 2772 and 13307, respectively. These challenges make the multi label classification task quite difficult and necessitate the incorporation of label dependencies along with employing robust learning approaches.

The standard approach to multi-label classification is Binary Relevance (BR), wherein the problem of multilabel classification for n classes is transformed into n binary classification problems \cite{DBLP:journals/corr/NamKGF13}. However, BR approach does not account for the interdependence of different class labels. The prevalent approaches to account for label dependencies include chain classification \cite{icml2010_DembczynskiCH10}, label power set of k-labels \cite{read2011classifier}, and modeling with recurrent neural network \cite{Yao2017LearningLabels}. The label power set approaches improve upon performance, but the modeling becomes intractable as the number of classes increase. Moreover, the use of recurrent neural network might not be able to model complex dependencies between class labels.

Another approach for multi label classification includes Pairwise Error (PWE) loss that inherently models label dependencies in the sense that it tries to maximize the margin between positive and negative labels within an example \cite{DBLP:journals/corr/GongJLTI13}\cite{Li2017ImprovingClassification}. Different variants of PWE loss are widely used for multi label classification in natural image classification and natural language processing tasks. However, the popular PWE losses such as hinge loss, margin loss are non-smooth and as such pose difficulty in optimization. Further, there exist empirical evidences that the ensemble approaches e.g. cascading, boosting, etc., demonstrate increased performance as compared to single classifiers. Although, the exact mechanism of how ensembling helps increase generalizability is still debatable. 

In the present work, we experiment with a series of deep learning methods for the diagnosis of chest diseases from the ChestX-ray14 dataset while systematically addressing the challenges mentioned in above paragraphs. To begin with, we train the popular DenseNet architecture \cite{HuangDenselyNetworks} with basic BR approach. Since the Chest X-ray14 dataset is quite large, we believe that training the neural network from scratch would allow for learning of more specific features corresponding to the underlying diseases as compared to fine-tuning a network trained on natural images. We also train the baseline architecture with PWE and compare the performance with cross entropy loss in BR method.
Further, we propose a novel cascading architecture that takes benefits from both ensemble and hard example mining techniques. We design the architecture such that it models the labels dependencies between classes inherently. The contributions of this paper can be summarized as:-

\begin{enumerate}
\item We train deep learning architecture from scratch for multi label classification of ChestX-ray14 dataset and achieve better results from baseline which used transfer learning.
\item We experiment two standard approaches to calculate loss for multi label classification: BR and PWE loss, and present comparison results.
\item We design a boosted cascade architecture that is specifically tailored for multi label classification task of type of ChestX-ray14 dataset. The proposed approach models complex dependencies between class labels and benefits from the training strategy of boosting methods to provide improved performance as compared to single classifiers trained using cross-entropy and PWE loss.
\end{enumerate}

\subsection{Related Work}
Multi label classification problem has been extensively explored for the tasks outside the domain of medical imaging e.g. object detection, text categorization. Last year Wang et. al. published a large dataset of chest X-ray images \cite{Wang2017ChestX-ray8:Diseases} that is a typical example of multi label classification problem in medical imaging domain. 

We have developed our method using DenseNet161 \cite{HuangDenselyNetworks} architecture and used the smoother version of pairwise error loss as described in \cite{Yeh2017LearningClassification}. Our cascading approach is inspired from basic Adaboost \cite{Schapire:1999:BIB:1624312.1624417} method that weighs individual examples while training different classifiers. The approach of forwarding output of preceding cascade level to next level bears similarity with the approach by Zeng et. al. \cite{Zeng2013Multi-stageDetection} used for object detection. Different from their work, we forward the outputs of all preceding cascade levels to next cascade level in the architecture.
.

\section{Methods}

We methodically experimented a series of approaches and evaluated their performance. We used DensNet161\cite{HuangDenselyNetworks} as baseline neural network architecture, due to its well known performance for a variety of datasets. The selection of the DenseNet architecture was additionally motivated by its efficiency in modeling with lesser number of parameters, which reduced the risk of over-fitting while still benefiting from deep architecture. Below, we briefly explain the approaches that we experimented:-
\subsection{Binary Relevance}
\begin{figure*}

\fbox{\includegraphics[width=\textwidth]{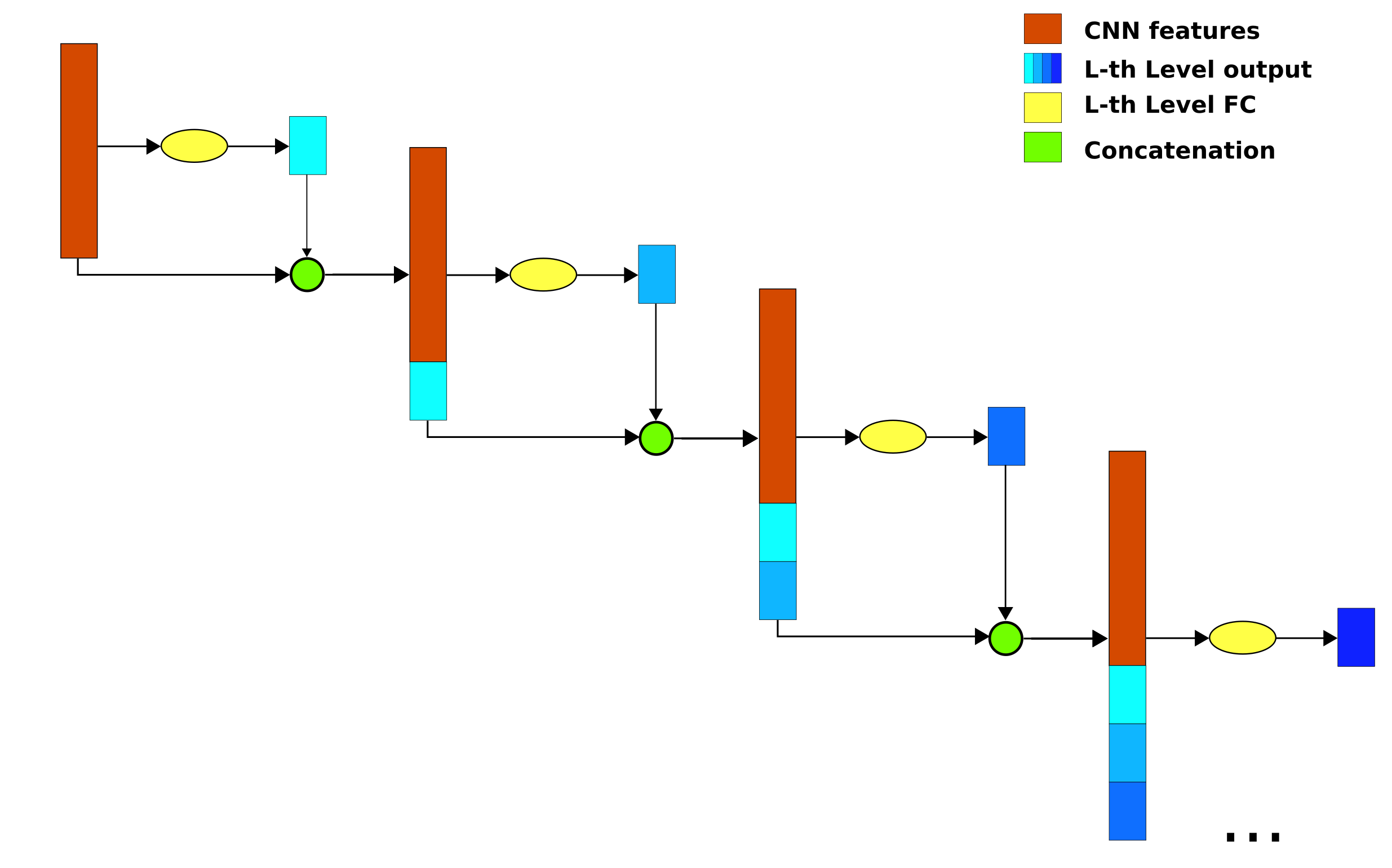}}

\caption{\scriptsize{Architecture for 4 levels of cascade}}
\label{fig}
\end{figure*}
For starter, we transformed the task as independent binary classification task for each class. Since the positive instances of any class are very less as compared to negative instances, we used weighted cross entropy loss by penalizing the errors on positive class instances by \textit{n/p}, where \textit{n} and \textit{p} correspond to frequencies of negative and positive instances, respectively. Moreover, we undersampled majority classes and oversampled minority classes to avoid bias due to imbalance in class occurrence and co-occurrence frequencies.
\subsection{PairWise Error (PWE) loss}
The major disadvantage of the BR approach is that it does not model the inter-class relations within an example. As an alternative, PWE loss maximizes the margin between the scores of each positive class to each negative class, and is based on simple logical reasoning that learning relations between pairs of each positive to negative class is sufficient to learn inter-class relations between all classes. We used smooth PWE loss \cite{Li2017ImprovingClassification} to separate probability scores of each positive class from each negative class within an example. The probability scores were computed as sigmoid output from the classifier layer. We used the sampling technique mentioned for BR approach to avoid adverse effects of class imbalance.
\begin{table*}[]
\centering
\begin{tabular}{|l|c|c|c|c|c|c|c|}
\hline
\multicolumn{1}{|c|}{} & \textbf{Wang et. al \cite{Wang2017ChestX-ray8:Diseases}} & \textbf{Yao et. al \cite{yao2017learning}} & \textbf{Rajpurkar et. al \cite{Rajpurkar2017CheXNet:Learning}} & \textbf{BR*} & \textbf{PWE*} & \textbf{C-BR*}   & \textbf{C-PWE*}  \\ \hline
\textbf{Atelectasis}   & 0.7158               & 0.772               & \textbf{0.8209}           & 0.7453      & 0.7158            & 0.7618          &   0.7433            \\ \hline
\textbf{Cardiomegaly}  & 0.8065               & 0.904               & 0.9048                    & 0.8947      & 0.8777            & \textbf{0.9133}  &   0.8930       \\ \hline
\textbf{Effusion}      & 0.7843               & 0.859               & \textbf{0.8831}           & 0.8396      & 0.8507            & 0.8635          &     0.8615   \\ \hline
\textbf{Infiltration}  & 0.6089               & 0.695               & \textbf{0.7204}           & 0.67        & 0.666             & 0.6923           &    0.6746        \\ \hline
\textbf{Mass}          & 0.7057               & 0.792               & \textbf{0.8618}           & 0.6964      & 0.7137            & 0.7502          &     0.7894   \\ \hline
\textbf{Nodule}        & 0.6706               & 0.717               & \textbf{0.7766}           & 0.6134      & 0.6023            & 0.6662          &     0.7035        \\ \hline
\textbf{Pneumonia}     & 0.6326               & 0.713               & \textbf{0.7632}                    & 0.55108     & 0.6073            & 0.7145 &    0.6378           \\ \hline
\textbf{Pneumothorax}  & 0.8055               & 0.841               & \textbf{0.8932}           & 0.8198      & 0.8408            & 0.8594          &      0.8531         \\ \hline
\textbf{Consolidation} & 0.7078               & 0.788               & \textbf{0.7939}                    & 0.7606      & 0.7324            & 0.7838 &       0.7604        \\ \hline
\textbf{Edema}         & 0.8345               & 0.882               & \textbf{0.8932}                    & 0.8669      & 0.8563            & 0.8880 &    0.8824    \\ \hline
\textbf{Emphysema}     & 0.8149               & 0.829               & \textbf{0.926}                     & 0.8474      & 0.8993            & 0.8982  & 0.9164 \\ \hline
\textbf{Fibrosis}      & 0.7688               & 0.767               & \textbf{0.8044}                    & 0.7236      & 0.7171            & 0.7559 & 0.7520 \\ \hline
\textbf{PT}            & 0.7082               & 0.765               & \textbf{0.8138}                    & 0.7405      & 0.7388            & 0.7739 & 0.7566 \\ \hline
\textbf{Hernia}        & 0.7667               & 0.914               & \textbf{0.9387}                    & 0.7122      & 0.898             & 0.8024 & 0.8636\\ \hline
\end{tabular}
\linebreak
\caption{\scriptsize{Comparison of AUCs of ROC curve for classification of diseases in ChestX-ray14 dataset. The experiments of this paper are marked with asterisk (*).}}
\label{result-table}
\end{table*}
\subsection{Boosted Cascade}

There exist empirical evidences that cascading multiple predictions using binary relevance improves performance of multi-label classification. Following this hypothesis, we trained a cascaded network to improve upon initial predictions from baseline networks using BR approach with cross-entropy loss, and PWE loss. We refer to these experiments as C-BR and C-PWE, respectively. Different from conventional cascading approaches, we designed our cascade network such that each succeeding level in the cascade network received predictions from all the preceding levels as input, as shown in Figure \ref{fig}. This allowed the different levels in the cascade network model non-linear dependencies between each class along with learning from mistakes of all preceding levels.

Conventionally, the different levels in the cascade are trained from different splits of the original training set. This is motivated by standard ensemble methods, wherein an ensemble of many weak classifiers is trained, each of which are independently tuned for smaller datasets. The said approach, thus considerably reduces the amount of training data for individual classifiers. Therefore its applicability is limited to ensembles of simpler models e.g. decision trees, support vector machines etc., as neural networks are prone to over-fitting if trained with smaller dataset. Therefore we trained different levels of the cascade network with entire training set. Further, we employed weighted sampling of the data points according to their difficulty level such that each successive cascade level was trained more for the data points which were difficult to classify by previous cascade level. 

After each cascade level was trained, loss was computed for each training example, which gave an estimate of difficulty level of each data point according to the classifier at l-th level. The data points were then sampled according to probability \textit{p} for training next cascade level. The probability of selection for i-th data point $p_i$ was computed according to following equation:-
\begin{equation}
\label{eqn}
p_i = \frac{e^\frac{-iR}{N}}{\sum\limits_{n=1}^{n=N}e^\frac{-nR}{N}}
\end{equation}
Where, N is total number of training data points; n belongs to [1, N]; R is rate of probability decay (higher the R, higher is the selection probability of difficult data points over easier data points). In this way, every succeeding classifier in the cascade focuses on more and more difficult examples.

\section{Implementation}
We used ChestX-ray14 dataset published by Wang et. al. \cite{Wang2017ChestX-ray8:Diseases}, which  contained 112,110 frontal X-ray images of 30,805 unique patients. The dataset contained labeling for fourteen different thoracic diseases. Following the method used in original paper, we randomly separated 20\% dataset for testing and utilized the rest for training.

We used DenseNet161 architecture and modified the last fully connected (\textit{fc}) layer to contain twice the number of classes as output units. The output was reshaped to compute independent softmax activation for each class against background.  Finally, weighted cross-entropy loss was computed. 

We used 6-levels of cascading for both cross entropy and PWE loss. Each level in cascading composed of two \textit{fc} layers. We used RELU non-linearity and a dropout of 0.5 between fc layers. The inference from cascading was made by taking average of the predictions from each level.

The training for each approach was done using  Stochastic Gradient Descent (SGD) with learning rate 0.1 and momentum 0.9. The learning rate was reduced to 1/10\textsuperscript{th} of the original value after 1/3\textsuperscript{rd} and 2/3\textsuperscript{rd} of the training finished. The network weights were initialized using He norm initialization.

\section{Results and Discussion}
Table [\ref{result-table}] represents the Area Under the Curve (AUC) of the Receiver Operating Characteristic (ROC) curve of the binary classification for all the classes from all the experiments. Both  PWE loss and BR with cross entropy loss performed comparable. Whereas, the combination of boosted cascade approach gave increased performance as compared to single classifier for both the losses.
Wang et al. \cite{Wang2017ChestX-ray8:Diseases} have provided baseline classification results for the dataset. Our baseline models: BR with cross-entropy, and PWE loss outperformed the baseline in 9 and 10 classes, respectively. Further, the addition of boosted cascade approach with both the losses outperforms the baseline in all the classes by a large margin. 

To the best of our knowledge, two other groups have reported their results on the ChestX-ray14 dataset. Yao et al. \cite{yao2017learning} proposed an LSTM based approach to model the label dependencies and Rajpurkar et. al. \cite{Rajpurkar2017CheXNet:Learning} use 121-layer convolutional neural network for classification. Our best performing model has better AUCs in 7 of the 14 classes as compared to Yao et al. and 1 of the 14 classes as compared to Rajpurkar et. al. \cite{Rajpurkar2017CheXNet:Learning} and comparable AUCs in rest of the classes. The performance comparison of our models with the baseline and other approaches is tabulated in Table [\ref{result-table}] itself.
\section{Conclusion}
We experiment a set of deep learning methods for the multi label classification of ChestX-ray14 dataset and provide results comparable to the state-of-the-art. We provide comparison results for cross entropy and pairwise error loss for the task of multi label classification of the dataset. Further, we implement a cascade network that improves upon the performance of deep learning models along with modeling label dependencies. In summary, the present work provides optimistic results for the automatic diagnosis of thoracic diseases. However, future work related to disease localization and improvement of classification performance is suggested.

{\small
\bibliographystyle{ieee}
\bibliography{egbib}
}

\end{document}